# Overcoming Imbalanced Safety Data Using Extended Accident Triangle

Kailai Sun[1], Tianxiang Lan[1], Yang Miang Goh[1*], and Yueng-Hsiang Huang[2]

[1]Department of the Built Environment, College of Design and Engineering, National University of Singapore, 117566, Singapore.

[2]Oregon Institute of Occupational Health Sciences, Oregon Health & Science University, 97239, United States.

*Corresponding author(s). E-mail(s): bdggym@nus.edu.sg;

Contributing authors: skl23@nus.edu.sg; tianxiang.lan@u.nus.edu; huangyu@ohsu.edu

## Abstract

There is growing interest in using safety analytics and machine learning to support the prevention of workplace incidents, especially in high-risk industries like construction and trucking. Although existing safety analytics studies have made remarkable progress, they suffer from imbalanced datasets, a common problem in safety analytics, resulting in prediction inaccuracies. This can lead to management problems, e.g., incorrect resource allocation and improper interventions. To overcome the imbalanced data problem, we extend the theory of accident triangle to claim that the importance of data samples should be based on characteristics such as injury severity, accident frequency, and accident type. Thus, three oversampling methods are proposed based on assigning different weights to samples in the minority class. We find robust improvements among different machine learning algorithms. For the lack of open-source safety datasets, we are sharing three imbalanced datasets, e.g., a 9-year nationwide construction accident record dataset, and their corresponding codes.

Keywords: Safety analytics, Accident triangle, Machine learning, Construction accidents, Trucking accidents, Imbalanced data

# 1 Introduction

According to the International Labour Organization (ILO) (2023), in 2019, more than 395 million workers worldwide suffered non-fatal injury at work. In addition, 2.93 million workers lose their lives to work-related accidents and diseases, with about 330,000 of these being fatal incidents. An alarming number of nearly 330 thousand work-related accidents are recorded in 2019. Despite significant efforts by governments and industries worldwide to reduce and prevent these injuries and ill health, they remain a persistent issue. Given the profound operational disruptions these injuries and ill health cause to businesses and the financial strain on societies, there is an increasing demand for data-driven approaches to support safety management, especially in high-risk industries like construction and trucking.

Safety analytics (Ezerins et al., 2022) is an emerging field of safety science that uses advanced quantitative methods to prevent workplace accidents and ill health. Safety analytics can identify the causes and drivers of workplace safety and health (WSH) incidents, highlighting high-risk activities to facilitate targeted interventions.

With the rapid development of safety analytics and artificial intelligence, machine learning (ML)-based analysis is gradually becoming an important tool to analyse and prevent workplace incidents (Sarkar & Maiti, 2020). For example, ML-based safety analytics were used to predict construction accident occurrence (Ghodrati et al., 2018) and severity (Hallowell et al., 2017), and identify safety management variables that influence these outcomes (Li et al., 2020). However, ML-based safety analytics studies frequently meet the imbalanced dataset problem (Jeong et al., 2018; Zhu et al., 2018) where one class (e.g., accident or failure occurrence) is significantly underrepresented compared to another (e.g., normal operations). A notable concern regarding the reliability of ML performance arises from imbalanced historical data (Ding & Xing, 2020). The performance of an ML classifier trained on an imbalanced dataset will exhibit a bias towards the majority class, which results in long-term application difficulties, e.g., unsafe predictions and improper interventions. The bias in models created using imbalanced dataset can have significant consequences in different subfields of safety analytics, including severity prediction of construction accidents (Koc & Gurgun, 2022; Luo et al., 2023), occurrence prediction of highway-rail grade crossings accidents (Gao et al., 2021), occurrence prediction of road crash accidents (Elamrani et al., 2024), etc.

Existing safety analytics studies usually use sampling methods to alleviate the imbalanced dataset problem, including undersampling (Gao et al., 2021; Kang & Ryu, 2019) and oversampling (e.g., Random Oversampling (ROS) (Gao et al., 2021; Koc & Gurgun, 2022; Schlögl, 2020)), Synthetic Minority Over-Sampling Technique (SMOTE) (Jeong et al., 2018; Luo et al., 2023; Peng et al., 2020), and Adaptive Synthetic Sampling (ADASYN)). However, there are limitations in applying these methods. Undersampling methods can lead to information loss and degrade model performance. In particular, if there is less data in the minority class, undersampling may lead to model underfitting. Since ROS involves replicating existing samples (Zheng et al., 2015), it often results in overfitting (Kotsiantis et al., 2006). SMOTE and ADASYN increase the risk of an overlapping problem by introducing noise (Ibrahim, 2021; Maciejewski & Stefanowski, 2011).

Rethinking the imbalanced data problem, we find that most existing safety analytics studies use the same data weights in sampling (Gao et al., 2021; Koc & Gurgun, 2022; Luo et al., 2023). Directly using these basic sampling methods means each minority class sample is treated as having the same weight. This assumption cannot hold in real-life scenarios. Different accidents have different characteristics and frequency of occurrence, and giving each accident the same weight cannot be the most appropriate approach. Previous work did not consider this.

The accident triangle is a potential solution, which suggests that there is a pattern in the ratio among different accident types. The original model by H.W. Heinrich (Anderson & Denkl, 2010; Hawkins & Fuller, 1996) suggested a 1-29-300 ratio among the frequency of major, minor, and no-injury incidents. Many studies (Bellamy, 2015; Gallivan et al., 2008) have proved that the ratio is not static but changes based on situation. Furthermore, the accident triangle is not concerned with employees' accident frequency and other accident characteristics, which are critical safety variables. Nevertheless, it is intuitive to assume that there will be fewer employees with a higher number of accidents and more employees with few or no accidents. Similarly, we expect different frequencies for different accident types in different industries (Mauro et al., 2018), e.g., more falling from height accidents in construction, and more vehicular accidents in transport. This study will exploit this natural extension of the accident triangle model (the extended accident triangle model) to guide the selection of sampling weights for predictive safety analytics based on accident characteristics, e.g., accident severity, employee accident frequency, and accident type.

The underlying assumption of the accident triangle is that there are common underlying causes of accidents of different severity. In a similar vein, there are also likely common underlying causes that contribute to workers' accident frequency, e.g., their risk tolerance (Haas et al., 2020), job satisfaction (Khoshakhlagh et al., 2021; López-García et al., 2019), work environment, and job type. These causes are, in turn, affected by systemic factors such as the organisation's safety climate (Bhandari & Hallowell, 2022), workload and job role clarity (Krueger et al., 2002). Such systemic factors may also be the common underlying causes across accident types with higher fatality risk or lower fatality risk. Thus, it is reasonable to account for the ratio of accident types and individuals with different characteristics during sampling by assigning different weights to samples in the minority class.

Besides, within the context of sampling weights, there are at least two possible interpretations of the extended accident triangle. First, higher weights can be assigned to higher frequency accidents or higher severity accidents. Major accidents are rarer and are of greater managerial significance because they may have stronger impacts, which are also more likely to be irreversible (e.g., permanent disability). Second, the extended accident triangle model also suggests that if the number of minor accidents is decreased, a corresponding fall in the number of major accidents will occur. This means we may assign higher weights to minor accidents as well. The key point is that the importance of each sample in the minority class may be treated differently based on a variable identified through domain knowledge. A weight may then be assigned to each sample to adjust its possibility of being utilized in oversampling.

Inspired by the concept of extended accident triangle, we claim: when using ML-based methods on safety data with different accident severity or frequency, each data point should *not* be treated as equally important. From a predictive safety analytics perspective, we ask: can the proportions of different incident characteristics (e.g., severity, frequency, and type) be used as a basis for assigning weights to different data points when applying oversampling methods to imbalanced safety datasets? This study gives us a confirmatory answer. Doing so allows us to introduce domain knowledge (the extended accident triangle) into existing sampling methods to improve ML models' predictive performance. In fact, prior or domain knowledge plays a crucial role in addressing the limitations of ML models, including data dependence, generalization capabilities, and adherence to constraints (Rath & Condurache, 2024; Xu et al., 2024). Inserting prior knowledge into ML networks is a hot topic in recent years (Dash et al., 2022), which aims to enhance existing ML models with specific domain knowledge (Tran & d'Avila Garcez, 2018; Weinberger et al., n.d.). In this work, we introduce safety prior knowledge (the extended accident triangle) at the data level to overcome the imbalanced data problem, improving ML models' predictive performance in safety analytics.

Previous studies have not explored the implications of the accident triangle and how to interpret it to facilitate the selection of sampling weights. The lack of attention on this issue could be due to the lack of publicly available safety datasets. Safety datasets can potentially reveal safety problems in different organizations. Thus, most researchers avoid sharing their data publicly. However, there are ways to share safety data without identifying the organizations and their safety issues. These datasets can be used to replicate and evaluate safety analytics methods, which are important in pushing the scientific frontier of the area.

In summary, this study addresses the following research challenges in safety analytics:

- Existing ML-based methods in safety analytics struggle to resolve the imbalanced data problem in safety datasets, resulting in poor performance.
- The famous accident triangle model focuses on the accident severity, but it does not include other accident characteristics, e.g., employee accident frequency, and accident type.
- ML-based methods in safety analytics require a lot of training data, but publicly available large and diverse tabular accident datasets, including construction accidents, are not common.

To address the above challenges, we introduce domain knowledge (the extended accident triangle) to adjust the weight of data samples to improve existing ML-based methods used in safety analytics. By presenting a general insight, we try to resolve a fundamental problem in safety analytics. This work has the following research contributions.

- We extend the accident triangle to cover employee's accident frequency and types during oversampling.
- Our insight is general, and thus, we apply our insight to three mainstream oversampling methods and proposed three improved oversampling methods. We tested them on different machine learning (ML) (including deep learning (DL)) methods.
- With more effective sampling, we demonstrate our methods on the three imbalanced datasets with superior performance.
- Three imbalanced safety datasets are established and shared: a 9-year nationwide construction accident dataset in Singapore, an accident and safety management dataset in a major development project in Singapore, and a US truck driver safety climate survey dataset. All three datasets are made publicly available.
- We provide the recommended accident-ratio weights for researchers and safety data to train their ML models. The code is available at https://github.com/NUS-DBE/accident.

# 2 Methods

With reference to Figure 1, in this study the data is captured from three different sources (discussed in Section 2.1). For each dataset, data points are resampled based on the concept of the Extended Accident Triangle to overcome data imbalance problems, and ML (including DL (e.g., neural network)) methods are trained to predict accidents outcomes.

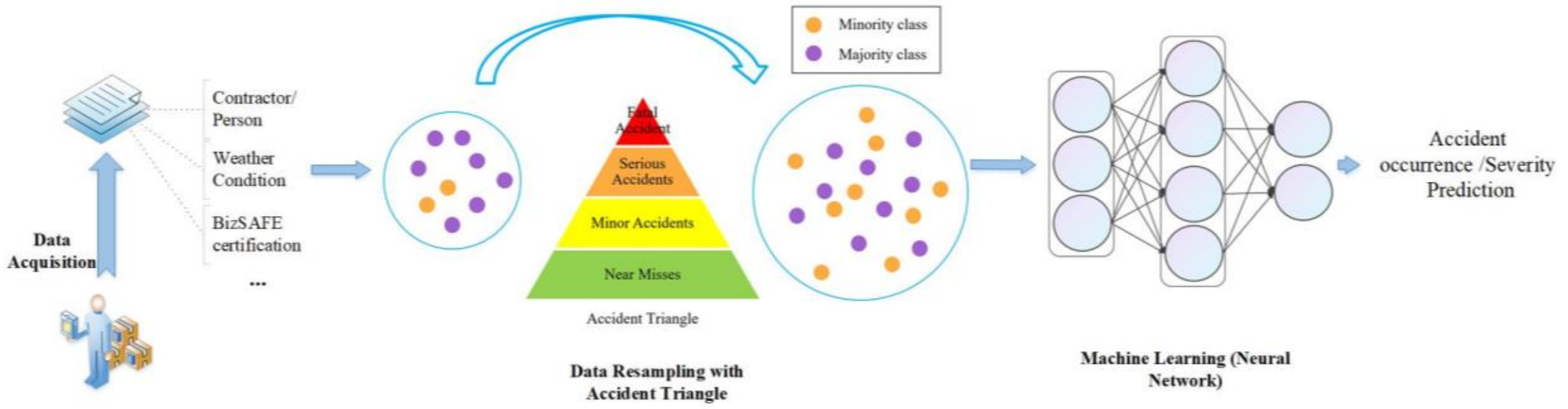

*Figure 1 The overall workflow for our proposed methods*

Based on this workflow, we divide this section into the following three parts. ML-based safety analytics require a lot of training data. Thus, we first establish and share three diverse accident datasets in Section 2.1. Then, we introduce the details of our proposed extended accident-triangle methods for safety analytics in Section 2.2. Finally, we train ML (including DL) methods to predict accidents in Section 2.3.

## 2.1 Accident data

ML and DL require a lot of training data. Unlike computer vision and natural language processing, publicly available tabular accident datasets are not common. This brings many challenges in advancing safety analytics, in particular, predictive analytics and development of safety leading indicators. Thus, we established three diverse accident datasets.

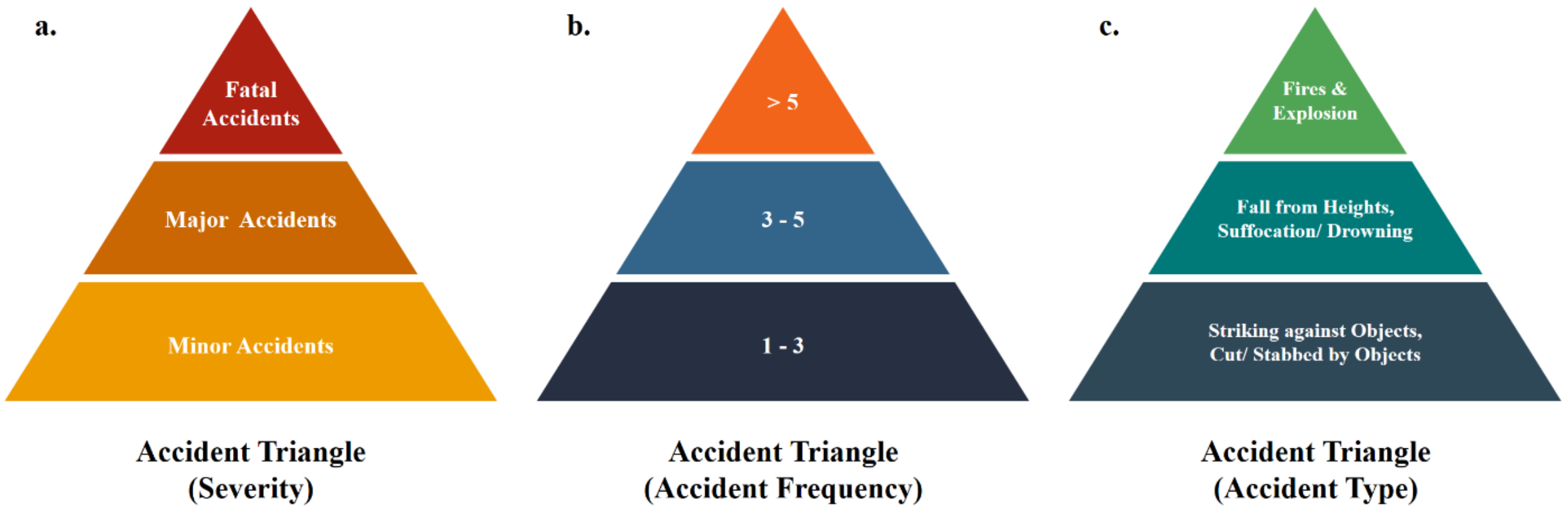

*Figure 2 Extended accident triangles. (a) Initially introduced in 1931 by Herbert William Heinrich (Heinrich & others, 1941), the theory suggests a relationship among the number of no-injuries, minor, and major accidents. Subsequent studies separated "fatalities" from major accidents, similar to the accident triangle in (a). In this study, we extend the accident triangle concept to the ratio across employee's accident frequency (b) and different accident types (c) (e.g., falling from height, struck by moving objects, and exposure to chemicals (Ministry of Manpower, 2024)). The contents of the accident triangles are for illustration only and may differ from dataset to dataset.*

The first dataset is based on the accident and operational record of a major development project in Singapore (Safety Management Dataset, SMD). It captures the safety management records of subcontractors for each month from July 2022 to June 2023, including variables such as inspection results, safety management system scores and number of accidents. This dataset is imbalanced because in each month, most sub-contractors have no accidents, and only a minority of them have accidents. The ratio between data points with accidents and those without is around 1:7.

The second dataset is based on the results of a safety climate survey among truck drivers in the United States (Safety Climate Dataset, SCD). SCD was set up to evaluate the relationship between trucking accidents and safety climate. This dataset is also imbalanced because most respondents reported no accidents for the past year, with a minority of them reporting one or more accidents. The ratio of respondents with and without accidents is around 1:3.

The third dataset is based on Singapore 9-year nationwide construction accident record (National Safety Dataset, NSD). NSD contains over 10,000 injury records including information such as accident agency, worker age, and weather conditions. The dataset is imbalanced because most accidents result in minor injuries, and the ratio between major and minor injuries is around 1:18.

### 2.1.1 SMD Dataset

SMD was established to evaluate the safety risk of construction sub-contractors. It contains records of sub-contractors' safety inspections, safety management scores and so on for each month. The date ranges from July 2022 to June 2023. SMD contains 998 samples: 878 samples have no accidents, while 120 samples have accidents. Each sample has 14 features as shown in Table 1. The accidents are categorised into three levels in order of severity (Figure 2(a)):

L1: Fatal Accidents

L2: Major Accidents: Workplace injury, dangerous occurrence, aerodrome incident, occupational diseases

L3: Minor Accidents: Property and services damage

Similar to mainstream accident prediction studies (Luo et al., 2023; Meng et al., 2022; Peng et al., 2020; Poh et al., 2018; Sarkar & Maiti, 2020), we consider the prediction of accidents occurrence a classification task. The samples with accidents were treated as a class, and those without another class. SMD is imbalanced. The class unbalance ratio is 120/878≈1/7. We randomly shuffled and split the dataset with a train-to-test ratio of 7:3. After splitting, we checked the feature range of the training and testing sets and found that they are similar. This ensures fair evaluation and other modelling issues. We first train ML/DL methods on the training set and then evaluate them on the testing set. When using EAT-based oversampling, samples with L2 accidents were assigned weight $\alpha$ and those with L3 accidents were assigned weight $\beta$. There are no L1 accidents recorded in the dataset.

*Table 1 Safety management dataset (SMD) description*

| Variables | Range | Description |
|---|---|---|
| Sub-contractor | - | The identity of the sub-contractor. |
| Main contractor | - | The identity of the main contractor that is the client of the sub-contractor. |
| Package | - | The work package the sub-contractor belongs to. Each package contains different ranges of work activities and may thus entail different levels or types of risks. |
| BizSAFE certification | 1-6 | BizSAFE is a national workplace safety and health certification programme. There are six levels of certification representing the firm's workplace safety and health management capacity. |
| Average number of inspections | 0-146.67 | The moving average of the number of safety inspections received by the sub-contractor for the past three months. |
| Months inspected | 0-3 | Out of the past three months, in how many months did the sub-contractor receive at least one safety inspection. |
| Inspection risk score | 5-25 | Risk scores assigned based on the variable "Months inspected" as proposed by the developer. 20, 15, 10, 5 scores are given to sub-contractors that were inspected at least once in 3, 2, 1, 0 of the past three months respectively. |
| Sum of observations per Inspection | 0-1 | Unweighted sum of number of observations recorded in inspections divided by the number of inspections over the past three months. "Observations" refer to either areas of improvement in safety practices (normal observations), or non-compliance with legal/developer's requirements and actions/conditions posing immediate danger to personnel (significant observations). |
| Weighted sum of observations per Inspection | 0-2 | Weighted sum of number of observations recorded in inspections divided by the number of inspections over the past three months. Significant observations are given twice the weight of normal observations. |
| Sub-contractor Safety Performance Score | 0-96 | The developer requires all operating sub-contractors' safety management system be assessed by the main contractor monthly. The scoring method is created by the developer and is based on assessing the sub-contractor's standard safety management procedures and their implementation. The maximal score is 100. |
| Main Contractor Site Safety Performance Assessment (SSPA) Score | 56-96 | The SSPA scoring method is created by the developer and is based on assessing the main contractor's standard safety management procedures and their implementation. The scoring is conducted monthly by the developer on the main contractors who are the clients of the sub-contractors. The maximal score is 100. |
| Main contractor Construction Safety Audit Scoring System (ConSASS) section 1 score | 90-100 | ConSASS is a national safety management audit checklist focusing on companies' safety management policies, resources, competency, contractor management, corrective actions to incidents and violations and continous improvement. The scoring is con- ducted twice yearly by third-party auditors on the main contractors who are the clients of the sub-contractors. Questions in the checklist are categorised into sections ("band"). This feature focuses on the section evaluating the safety management procedure. The maximal score is 100. |
| Main contractor Construction Safety Audit Scoring System (ConSASS) section 2 score | 63-100 | This feature focuses on the section evaluating the implementation of the safety management procedure on site. The maximal score is 100. |
| Main contractor CultureSAFE score | 2.1-4.2 | CultureSAFE is a national safety culture assessment tool. The scoring is conducted twice yearly through the main contractor's self-assessment. The maximal score is 5. |

### 2.1.2 SCD Dataset

SCD consists of safety climate survey data among truck drivers collected by Huang et al. (2013). The survey was distributed to eight trucking companies in the U.S. via a web-based portal and conventional pen-and-paper method from 2011 to 2013. The dataset has a total of 7,474 entries. 366 incomplete data points with more than 10% missing values were removed because the dataset is large enough to allow the removal of those data points with marginal loss in power. The missing values of data points with less than 10% missing values were imputed using the median replacement function suggested by Lynch (2007). The survey asked respondents to rate their perceptions of safety in their company using a five-point Likert scale (1 = Strongly Disagree, and 5 = Strongly Agree). The SCD includes 42 safety climate perception items measured on a five-point Likert scale collected by Huang et al. (2013), the details of which are shown in Table 2. The survey also asked respondents the number of accidents they suffered in the past year. Whether the respondent had accidents in the past year may be taken to reflect how accident-prone each respondent is and, thus, be used as the target in the dataset. The data is imbalanced, as most respondents reported no accidents. Specifically, 5135 reported no accidents, 1499 reported suffering from one accident, and 475 reported suffering from more than one accident. The ratio of respondents with and without an accident is 1974/5135≈1/3. We coded respondents into three levels: those without accidents (least accident-prone), those reporting one accident (moderately accident-prone) and those reporting more than one accident (most accident-prone). We randomly shuffled and split the dataset with a train-to-test ratio of 7:3 and developed supervised learning models to classify whether one had accidents in the past year. The feature range of the training and testing sets are also found to be similar, assuring fair evaluation. We first train ML/DL methods on the training set and then evaluate them on the testing set. When using EAT-based oversampling, samples recording more than one accident were assigned weight $\alpha$, and those recording only one accident were assigned weight $\beta$.

*Table 2 Safety climate dataset (SCD) description*

| S/N | Variable | Description | |
|---|---|---|---|
| 0 | Accident | Driver's self-reported number of accidents in the previous year. 0 = no accident, 1 = one accident 2 = more than one accident. | |
| **Below are items measuring truck drivers' safety climate perception. All items are measured in a 1-5 Likert scale.** | | | |
| **S/N** | **Item details** | **S/N** | **Item details** |
| 1 | Uses any available information to improve existing safety rules | 22 | Provides me with feedback to improve my safety performance |
| 2 | Tries to continually improve safety levels in each department | 23 | Respects me as a professional driver |
| 3 | Invests a lot in safety training for workers | 24 | Frequently talks about safety issues throughout the work week |
| 4 | Creates programs to improve drivers' health and wellness (e.g., diet, exercise) | 25 | Discusses with us how to improve safety |
| 5 | Listens carefully to our ideas about improving safety | 26 | Uses explanations (not just compliance) to get us to act safely |
| 6 | Cares more about my safety than on-time delivery | 27 | Is supportive if I ask for help with personal problems or issue |
| 7 | Allows drivers to change their schedules when they are getting too tired | 28 | Is an effective mediator/trouble-shooter between the customer and me |
| 8 | Provides enough hands-on training to help new drivers be safe | 29 | Is strict about working safely even when we are tired or stressed |
| 9 | Gives safety a higher priority compared to other truck companies | 30 | Gives higher priority to my safety than on-time delivery |
| 10 | Reacts quickly to solve the problem when told about safety concerns | 31 | Would like me to take care of serious equipment problems first before delivering |
| 11 | Is strict about working safely when delivery falls behind schedule | 32 | Gives me the freedom to change my schedule when I see safety problems |
| 12 | Gives drivers enough time to deliver loads safely | 33 | Makes me feel like I'm bothering him/her when I call |
| 13 | Fixes truck/equipment problems in a timely manner | 34 | Encourages us to go faster when deadheading (going for a new load) |
| 14 | Will overlook log discrepancies if I deliver on time | 35 | Expects me to sometimes bend driving safety rules for important customers |
| 15 | Makes it clear that, regardless of safety, I must pick up/deliver on time | 36 | Sometimes turns a blind eye with rules when deliveries fall behind schedule |
| 16 | Expects me to sometimes bend safety rules for important customers | 37 | Pushes me to keep driving even when I call in to say I feel too sick or tired |
| 17 | Turns a blind eye when we use hand-held cell phones while driving | 38 | Expects me to answer the cell phone even while I'm driving |
| 18 | Assigns too many drivers to each supervisor, making it hard for us to get help | 39 | Stops talking to me on the phone if he/she hears that I am driving |
| 19 | Hires supervisors who don't care about drivers | 40 | Turns a blind eye when we use hand-held cell phones while driving |
| 20 | Turns a blind eye when a supervisor bends some safety rules | 41 | Is an effective mediator / trouble-shooter between management and me |
| 21 | Compliments employees who pay special attention to safety | 42 | Regularly asks me if I have had enough sleep |

### 2.1.3 NSD Dataset

NSD consists of the record of Singapore's nationwide construction accident data from 2014 to 2022, with a total of 10757 injury cases. The record includes the nature and agency of the accident that caused the injury, the nature of the injury and the worker's age and gender. It also includes the rainfall amount (in millimetres) and duration (in minutes) during the same hour as the accident, as well as during the hour before, two, three, four, five and six hours before the accident, as recorded by the weather station nearest to the accident location. The details of the dataset are shown in Table 3. These variables were used as the features in the model. The dataset also includes the injury severity (major or minor), which served as the target variable. As explained earlier, this dataset is also imbalanced, with 575 major injuries and 10182 minor injuries, a class imbalance ratio of 575/10182≈1/18. We randomly shuffled and split the dataset with a train-to-test ratio of 7:3. We checked the feature range of the training and testing sets and they are found to be similar. We first train ML/DL methods on the training set and then evaluate them on the testing set. According to the 2023 Workplace Safety and Health Report (Ministry of Manpower, 2024), we assigned weights to different samples based on the accident types with higher or lower fatality risk. Table 4 shows the accident types and the weights assigned to samples having such accidents.

*Table 3 National safety dataset (NSD) description*

| **Variables** | **Range** | **Description** |
|---|---|---|
| Severity | 0-1 | Severity of the workplace injury recorded. 0 means minor injury, 1 means major injury. |
| Agency | - | The agency of the accident (e.g., vehicle, flying fragments) |
| Age | 15-84 | The injured worker's age |
| Gender | M, F | The injured worker's gender |
| Nearest weather station | - | The weather station nearest to where the accident happened. Used as a proxy for the location of the accident. |
| Rainfall amount | 0-50.2 | The amount of rainfall (mm) recorded by the nearest weather station during the hour when the accident happened. For example, if the accident happened at 15:30, this variable registers the total rainfall between 15:00 to 16:00. |
| Rainfall duration | 0-60 | The number of minutes that was raining recorded by the nearest weather station during the hour when the accident happened. For example, if the accident happened at 15:30, this variable registers the total duration of rain between 15:00 to 16:00 |
| Rainfall amount 1-hour prior | 0-62.8 | The amount of rainfall (mm) recorded by the nearest weather station one hour before the accident. For example, if the accident happened at 15:30, this variable registers the total rainfall between 14:00 to 15:00. |
| Rainfall duration | 0-60 | The number of minutes that was raining recorded by the nearest weather station one hour before the accident. For example, if the accident happened at 15:30, this variable registers the total duration of rain between 14:00 to 15:00. |
| Rainfall amount 2-hour prior | 0-43.8 | The amount of rainfall (mm) recorded by the nearest weather station the second last hour before the accident. For example, if the accident happened at 15:30, this variable registers the total rainfall between 13:00 to 14:00. |
| Rainfall duration | 0-60 | The number of minutes that was raining recorded by the nearest weather station the second last hour before the accident. For example, if the accident happened at 15:30, this variable registers the total duration of rain between 13:00 to 14:00. |
| Rainfall amount 3-hour prior | 0-48.6 | The amount of rainfall (mm) recorded by the nearest weather station the third last hour before the accident. For example, if the accident happened at 15:30, this variable registers the total rainfall between 12:00 to 13:00. |
| Rainfall duration | 0-60 | The number of minutes that was raining recorded by the nearest weather station the third last hour before the accident. For example, if the accident happened at 15:30, this variable registers the total duration of rain between 12:00 to 13:00. |
| Rainfall amount 4-hour prior | 0-45.6 | The amount of rainfall (mm) recorded by the nearest weather station the fourth last hour before the accident. For example, if the accident happened at 15:30, this variable registers the total rainfall between 11:00 to 12:00. |
| Rainfall duration | 0-60 | The number of minutes that was raining recorded by the nearest weather station the fourth last hour before the accident. For example, if the accident happened at 15:30, this variable registers the total duration of rain between 11:00 to 12:00. |
| Rainfall amount 5-hour prior | 0-52.6 | The amount of rainfall (mm) recorded by the nearest weather station the fifth last hour before the accident. For example, if the accident happened at 15:30, this variable registers the total rainfall between 10:00 to 11:00. |
| Rainfall duration | 0-60 | The number of minutes that was raining recorded by the nearest weather station the fifth last hour before the accident. For example, if the accident happened at 15:30, this variable registers the total duration of rain between 10:00 to 11:00. |
| Rainfall amount 6-hour prior | 0-45.6 | The amount of rainfall (mm) recorded by the nearest weather station the sixth last hour before the accident. For example, if the accident happened at 15:30, this variable registers the total rainfall between 9:00 to 10:00. |
| Rainfall duration | 0-60 | The number of minutes that was raining recorded by the nearest weather station the sixth last hour before the accident. For example, if the accident happened at 15:30, this variable registers the total duration of rain between 9:00 to 10:00. |

*Table 4 The fatality risk of accident types in NSD*

| **Accident type** | **Weight** | **Fatality risk** | **No. of samples in the original dataset** |
|---|---|---|---|
| Struck by Moving Objects | α | High | 2347 |
| Caught in/between Objects | | | 1053 |
| Collapse/Failure of Structures | | | 43 |
| Fires & Explosion | | | 46 |
| Falls from Heights | | | 1320 |
| Struck by Falling Objects | | | 1365 |
| Striking against Objects | | | 504 |
| Over-exertion/Strenuous Movements | β | Low | 342 |
| Exposure to Hazardous Substances | | | 141 |
| Others | | | 91 |
| Exposure to Extreme Temperatures | | | 70 |
| Exposure to Biological Materials | | | 62 |
| Stepping on Objects | | | 60 |
| Others-Traffic Accident | | | 51 |
| Slips, Trips & Falls | | | 2020 |
| Cut/Stabbed by Objects | | | 1198 |
| Physical Assault | | | 25 |
| Exposure to Electric current | | | 19 |

In summary, three imbalanced datasets, representative of different societal levels - Company, Industry, and National - are documented from real-world scenarios. The three datasets vary in their data types, encompassing mandatory accident reports, company safety management data, and safety climate perception surveys.

More importantly, to promote the use of advanced safety analytics, we are sharing all three datasets. To protect the privacy of stakeholders, we removed sensitive details, e.g., time (for accidents), gender and age (for survey data).

## 2.2 Extended accident triangle (EAT) sampling methods

We will introduce the details of our proposed extended accident triangle (EAT) methods in the following subsections.

### 2.2.1 EAT-ROS

To overcome the imbalance data issues, many accident analysis studies (Gao et al., 2021; Koc & Gurgun, 2022; Schlögl, 2020) are based on a naive oversampling method: random oversampling (ROS). As one of the simplest techniques, ROS usually aims to expand the dataset by copying samples from the minority class set. However, there are many limitations.

For example, directly applying ROS in safety analytics does not consider the specific knowledge in the safety field, and it is difficult to achieve optimal solutions.

In this subsection, we present the extended accident triangle-Random Oversampling (EAT-ROS) method, an improved oversampling method. First, we determine the EAT weight for each sample by assigning a weight $\alpha$ if the sample contains a major accident and assigning a weight $\beta$ if the sample contains a minor accident. We assign no weight (0) if the sample contains no injury. Second, we randomly choose a sample based on the above EAT weights to be duplicated to augment the minority class. This effectively adjusts the sampling weights according to the severity of accidents.

The detailed pseudo-code of EAT-ROS is shown in Algorithm 1. Given an imbalanced accident dataset which contains minority classes and majority classes, we expect to balance this accident dataset by oversampling. Given the set of minority class samples $D$, we compute the EAT weights of minority class samples $W$ according to the above weight assignment. We normalize the $W$ as a probability density function (pdf). Then, we determine how many minority class samples to generate $N$, e.g., the number of majority class samples minus the number of minority class samples. Furthermore, we repeat the sample $x$ based on $W$. Last, we add the set of generated samples $S$ into this minority class data to achieve a balanced dataset.

**Algorithm 1:** Extended-Accident-Triangle Random Oversampling (EAT-ROS)

**Input:** Set of minority class samples $D$;
Extended-Accident-Triangle weights of minority class samples $W$;
Number of samples to generate $N$.
**Output:** Set of generated samples $S$

Normalize $W \leftarrow \frac{W}{\sum W}$ ;
Initialize an empty set $S$;
**while** $N \neq 0$ **do**
    $x \leftarrow$ Choose a random sample from $D$ based on weights $W$;
    $S \leftarrow S \cup \{x\}$;
    $N \leftarrow N - 1$;
**end**
**return** $S$

### 2.2.2 EAT-SMOTE

SMOTE is an oversampling approach aiming at addressing the class imbalance problem in datasets used for machine learning. When one class dominates over the others, the predictive performance, especially for the minority class, can be negatively impacted. SMOTE helps by generating synthetic samples in the feature space. SMOTE addresses the primary limitation of random oversampling (which can lead to overfitting) by generating synthetic samples rather than just duplicating existing ones. The key features of SMOTE include synthetic sample creation, interpolation and $k$ nearest neighbours.

Many accident analysis studies use SMOTE to address imbalanced problems (Jeong et al., 2018; Luo et al., 2023; Peng et al., 2020). EAT-SMOTE is an improved version of SMOTE by introducing the extended accident triangle in Algorithm 2. The EAT weights $W$ are obtained by similar assignment in Section 2.2.1. To synthesize $N$ new samples, the variable $i$ is set to the remainder of $N$ divided by the size of dataset $D$. This effectively gets an index of a sample from minority class $D$. Then, we compute the $k$ nearest neighbors ($K$) for each sample $D_i$. We choose a random sample $D_i$ from $K$, and synthesize new samples using $D_i + W_i^j \times (D_j - D_i)$. In standard SMOTE, $W_i^j$ will be randomly assigned from (0,1). Instead, our EAT-SMOTE assigns the $W_i^j$ from the random samples in Gaussian distribution $N(\frac{W_i}{W_i+W_j}, \sigma^2)$. It means that we use the EAT weights $W$ to synthesize new samples. In detail, we sample weight from the Gaussian distribution where the mean is $\frac{W_i}{W_i+W_j}$ and the variance is $\sigma^2$.

**Algorithm 2:** Extended-Accident-Triangle SMOTE ( EAT-SMOTE)

**Input:** Set of minority class samples $D$;
Number of nearest neighbors $k$;
Extended-accident-triangle weights of minority class samples $W$;
Number of samples to synthesize $N$.
**Output:** Set of synthetic samples $S$

Normalize $W \leftarrow \frac{W}{\sum W}$ ;
Initialize an empty set $S$;
**while** $N \neq 0$ **do**
  $i \leftarrow N \% |D|$;
  $K \leftarrow$ Compute the $k$ nearest neighbors in $D$ for each sample $D_i$;
  $D_j \leftarrow$ Choose a random sample from $K$;
  $W_i^j \leftarrow N(\frac{W_i}{W_i+W_j}, \sigma^2)$;
  $x \leftarrow D_i + W_i^j \times (D_j - D_i)$;
  $S \leftarrow S \cup \{x\}$;
  $N \leftarrow N - 1$;
**end**
**return** $S$

### 2.2.3 EAT-ADASYN

ADASYN is an algorithm designed to address the challenge of class imbalance in machine learning datasets. The key features of ADASYN include adaptive sampling, generating synthetic samples, and $k$ nearest neighbours. By focusing on generating synthetic samples in regions of the feature space where the minority class is underrepresented, ADASYN can lead to improved classification performance, especially in regions where the decision boundary is ambiguous.

Similarly, EAT-ADASYN is an improved version of ADASYN by introducing the extended accident triangle in Algorithm 3. As we know, the overall accident dataset can be divided into many classes, including minority classes. We choose a set of minority class samples $D$ to begin with the EAT-ADASYN. The EAT weights $W$ are obtained by similar assignment as in Sections 2.2.1 and 2.2.2. Then, we compute the $k$ nearest neighbors $(K)$ in $C \cup D$ for each sample $D_i$ in $D$. Note that the $K$ is different from SMOTE and EAT-SMOTE. For each sample $D_i$, to determine the number of samples to synthesize, we compute its density distribution $d_i$ individually. Then, we compute the number of synthetic samples for each sample $D_i$ as $\lfloor N \times \frac{d_i}{\sum d_i} + 0.5 \rfloor$. Next, we introduce how to synthesize new samples. Given the sample $D_i$, we choose a random sample $D_j$ from $K \cap D$. The minority class sample pair $(D_i, D_j)$ is constructed. Following the EAT-SMOTE, our EAT-ADASYN assigns the $W_i^j$ from the random samples in Gaussian distribution $N(\frac{W_i}{W_i + W_j}, \sigma^2)$. Lastly, we synthetize new samples using $D_i + W_i^j \times (D_j - D_i)$.

---

**Algorithm 3:** Extended-Accident-Triangle ADASYN ( EAT-ADASYN)

---

**Input:** Set of minority class samples $D$;
Set of other class samples $C$;
Number of nearest neighbors $k$;
Extended-accident-triangle weights of minority class samples $W$;
Number of samples to synthesize $N$.
**Output:** Set of Synthetic samples $S$

Normalize $W \leftarrow \frac{W}{\sum W}$ ;
Initialize an empty set $S$;
  $K \leftarrow$ Compute $k$ nearest neighbors in $C \cup D$ for each sample $D_i$ in $D$;
  $d_i \leftarrow \frac{|K \cap D|}{k}$.
  $g_i \leftarrow \lfloor N \times \frac{d_i}{\sum d_i} + 0.5 \rfloor$
  **for** *each sample* $D_i$ *in* $D$ **do**
    **for** $j = 1$ *to* $g_i$ **do**
      $D_j \leftarrow$ Choose a random sample from $K \cap D$
      $W_i^j \leftarrow N(\frac{W_i}{W_i + W_j}, \sigma^2)$;
      $x \leftarrow D_i + W_i^j \times (D_j - D_i)$;
      $S \leftarrow S \cup \{x\}$;
    **end**
**end**
**return** $S$

---

## 2.3 Machine learning methods

We apply different ML methods to test our proposed insight. Because machine learning methods (e.g., gaussian naive Bayes, logistic regression, and decision tree) have been well developed and used in a wide range of studies, we will focus on the introduction of the deep learning method we used.

The architecture of our Convolutional Neural Network (CNN) model comprises four convolutional layers, each followed by a max-pooling operation. The number of filters in successive layers is 16, 32, 64, and 128, with a consistent kernel size of 3, stride of 1, and padding of 1. While the first three layers employ a max-pooling kernel size of 2 and stride of 2, the fourth layer uses a pooling kernel size of 1 to retain feature map dimensions. Post convolution, the features are flattened and processed by a 256-neuron dense layer with ReLU activation. The final layer provides class predictions. For the training phase, we use the cross-entropy loss as our objective function:

$$l(\theta) = -\sum_{i=1}^{C} y_i \log(p_i),$$

where $C$ is the number of classes; $y_i$ is the class label; $p_i$ is the predicted probability of the observation being of class $i$. Then, the Adam optimizer, with a learning rate of 0.003, is chosen for its adaptability and efficiency in training deep neural networks. The batch size is set to 32. Our model is trained for 10 epochs, during which we monitor the validation accuracy to save the state of the best-performing model.

# 3 Evaluation metrics

Many studies formulate the accident prediction task as an accident classification task. For evaluation metrics, we use the balanced accuracy (BA) (Kelleher et al., 2020):

$$BA = \frac{1}{2}\left(\frac{TP}{TP + FN} + \frac{TN}{TN + FP}\right),$$

where $\frac{TP}{TP+FN}$ is the sensitivity (True Positive Rate, TPR), and $\frac{TN}{TN+FP}$is the specificity (True Negative Rate, TNR). TP is True Positive; TN is True Negative; FP is False Positive; FN is False Negative.

When a classifier demonstrates equal performance across both classes in a balanced dataset, the balanced accuracy aligns with the traditional accuracy measure, which is simply the ratio of correct predictions to the total number of predictions. However, in situations where the conventional accuracy appears deceptively high due to an imbalanced dataset, the balanced accuracy provides a more representative metric. If the classifier's success is solely because it exploits the imbalance in the test set, the balanced accuracy will appropriately adjust, often dropping to chance level. This refined explanation underscores the importance of balanced accuracy as a metric, especially in scenarios with class imbalances.

In particular, in Figure 3 (b) and (d), we use the increasing balanced accuracy (IBA) to highlight the increase of our insight:

$$IBA = BA_w - BA,$$

where $BA_w$ is the balanced accuracy with our EAT weights, and $BA$is the balanced accuracy without our EAT weights for each dataset and ML and DL method.

# 4 Results

We present the results of our experiments in this section (summarized in Figure 3). First, we present the effectiveness of our three oversampling methods on three predictive safety analytics tasks and datasets (SMD, SCD, NSD). Then, each task is performed using three to four state-of-the-art (SOTA) ML or DL algorithms to validate the robustness of our oversampling methods.

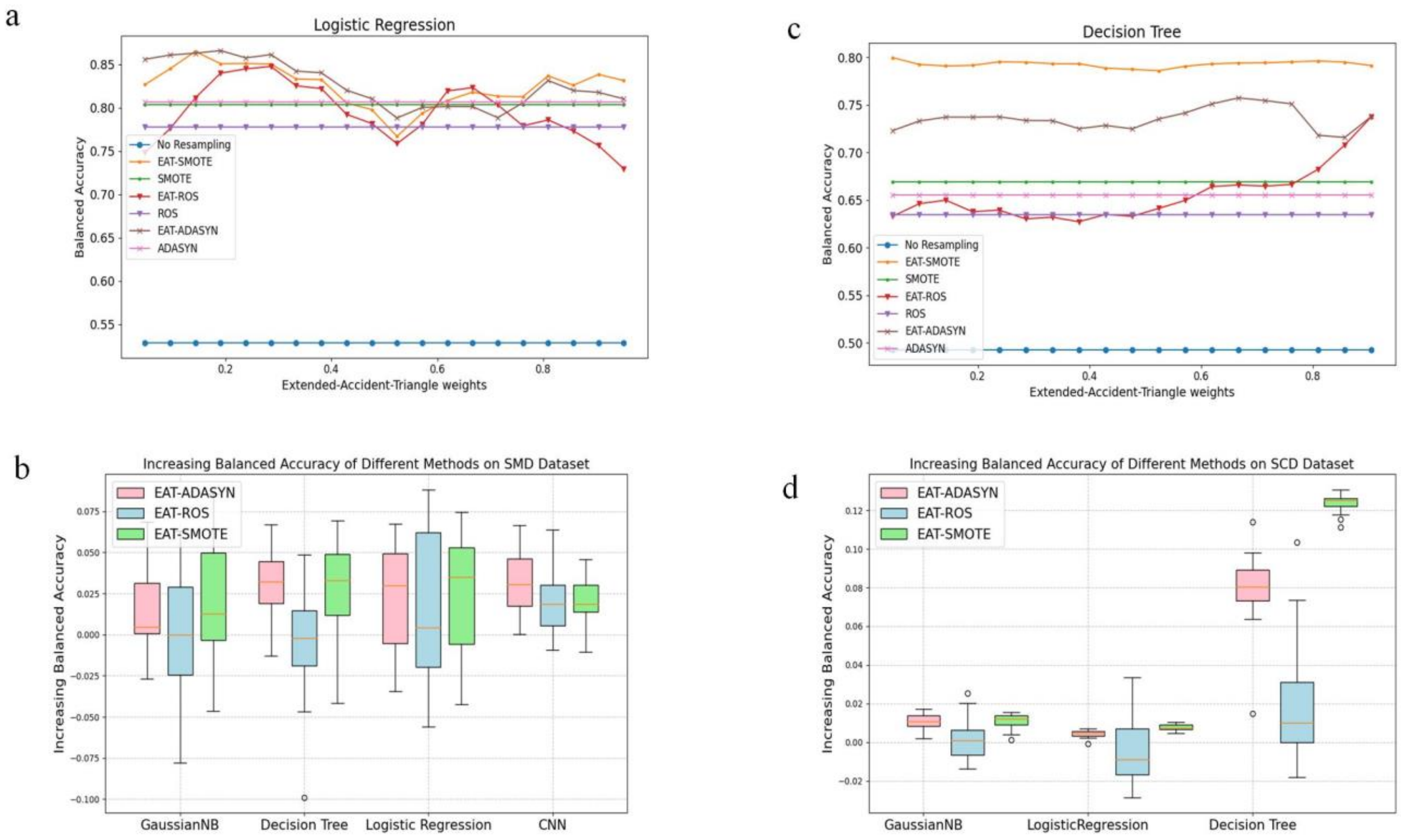

*Figure 3 Performance analysis of our methods on dataset SMD and SCD. (a) Comparison of extended-accident-triangle series methods against baselines on Dataset SMD using Logistic Regression. (b) The Increasing balanced accuracy of different methods on SMD dataset. (c) Comparison of extended-accident-triangle series methods against baselines on Dataset SCD using Decision Tree. (d) The Increasing balanced accuracy of different methods on SCD dataset. (Detailed charts are in Supplementary Material)*

## 4.1 Sub-contractor accident occurrence prediction: weight vs. no weight

To compare to the SOTA methods, we tested the balanced accuracy (defined in Section 3) of different machine learning (gaussian naive Bayes, logistic regression, decision tree) and deep learning (convolutional neural network) methods on dataset SMD in Figure 1, and Figures 1, 2, and 3 in the supplementary material. The safety analytics task is to predict whether a given subcontractor has accidents in a given month using the safety management and operation data of the same month. The 14 input variables of ML/DL methods are shown in Table.1 while the output variable of ML/DL methods is the accident occurrence.

In Figure 3 (b), we used a mainstream deep learning method (convolutional neural network) to predict accident occurrence. We trained ML models to predict the sub-contractor accident occurrence on dataset SMD. First, we obtain the source data. Second, we oversample the minority class of the datasets. Third, we formulate it as an ML or DL prediction task, with the accident occurrence as the label and others as the feature. After training, the ML or DL model can predict accident occurrence for given features. We fixed the training flow and compared the baseline oversampling methods (ROS, SMOTE and ADASYN) with our EAT-based methods (EAT-ROS, EAT-SMOTE and EAT-ADASYN).

For the EAT-based methods on the SMD dataset, two weights, $\alpha$ and $\beta$ were assigned to minority class samples with high-impact and low-impact accidents respectively, such that $\alpha + \beta = 1$. We tried 20 different combinations of $\alpha$ and $\beta$, with $\alpha$ increasing from 0 to 1 by increments of 0.048 and $\beta$ decreasing from 1 to 0 correspondingly. For better visualisation, we also plotted balanced accuracy against $\alpha$ value in Figure 3 (a).

Because the dataset is imbalanced (class unbalance ratio between samples with and without accidents:$\frac{120}{878} \approx \frac{1}{7}$), the baseline balanced accuracy without oversampling is low (blue line), at about 66.1%. We oversampled the minority class dataset to achieve the 1:1 balance. We retrained the ML model using the oversampled datasets. The accuracies improved significantly after standard oversampling: about 80.4% by SMOTE, about 77.8% by ROS, and about 80.7% by ADASYN. Nevertheless, in Figure 3 (a), the accuracies further improved after applying our oversampling methods: the highest accuracies are 87.8% by EAT-SMOTE; 86.6% by EAT-ROS; 87.4% by EAT-ADASYN. Our methods achieve a significant improvement of about 6.7-8.8%. It is noteworthy that EAT-SMOTE and EAT-ADASYN outperformed standard SMOTE and ADASYN on most $\alpha$ values tested and EAT-ROS outperformed standard random oversampling on most $\alpha$ values tested.

To make the comparison clearer, in Figure 3 (b), we show how much higher the balanced accuracy values are when using EAT-based oversampling methods compared to using standard oversampling methods. The positive average value in the boxplot shows that, on average, EAT-based oversampling methods resulted in higher balanced accuracy than other methods for all ML and DL algorithms tested. This demonstrates that our insight brings a stable positive increase compared to the baseline. It means that if we add extended-accident-triangle weights when oversampling, higher accuracy will be achieved with a high probability, compared with the no weight. In summary, from Figure 3 (a) and (b), the classification tasks conducted based on dataset oversampled using EAT-based methods performed better than using other methods (ROS, SMOTE, ADASYN), suggesting that EAT-based oversampling enhances the accuracy of ML and DL-based prediction of accident occurrence.

## 4.2 Employee accident occurrence prediction: weight vs. no weight

To evaluate the robustness of our methods, we conducted similar experiments on the SCD dataset. The safety analytics task is to predict whether a driver has had any accidents in the past year using 42 safety climate perception scores. The input variables of ML/DL methods are 42 safety climate perception scores in Table.2 while the output variable of ML/DL methods is the accident occurrence. This target helps the organization identify employees who are more likely to have accidents. The dataset is imbalanced (ratio between drivers with and without accidents$\frac{1974}{5134} \approx \frac{1}{3}$).

For the drivers with accidents for the past year, we split them into two types: the drivers who have had two or more accidents, and the drivers who have had one accident. For the EAT-based methods on the SCD dataset, two weights, $\alpha$ and $\beta$ were assigned to minority class samples with different drivers, such that $\alpha + \beta = 1$.

We tested the balanced accuracy of different machine learning methods (gaussian naive Bayes, logistic regression, decision tree) on SCD. In Figure 3 (c), and Figures 4 and 5 in the supplementary material, using standard oversampling methods, the three machine learning methods did not perform well. As observed in Figure 3 (c), our EAT-SMOTE and EAT-ADASYN outperformed standard SMOTE and ADASYN on all EAT weights. In addition, the positive average value in the boxplot shows that, on average, EAT-based oversampling methods always resulted in higher balanced accuracy than standard oversampling methods. In Figure 3 (d), all classification models tested yielded higher average balanced accuracy when run on datasets oversampled with EAT-based methods than with other methods. The differences in balanced accuracies when using the EAT-based method and baseline oversampling method were greater than 0 among different ML methods. It showed a significant increase in accuracy when using the decision tree algorithm. On the other hand, ROS (EAT-ROS) generates samples by random sampling instead of introducing new samples, so the variance of EAT-ROS in Figure 3 (d) was bigger than other EAT-based methods.

## 4.3 Accident injury severity prediction: weight vs. no weight

We applied our oversampling approach to another common safety analytics task, injury severity prediction. We obtained and cleaned the 9-year nationwide construction injury record dataset. In it, each injury is tagged to a weather station that is nearest to the location of the accident that caused the injury. The 18 input variables of ML/DL methods are shown in Table.3 while the output variable of ML is the accident injury severity .Three-year injury heatmaps of Singapore are visualized inFigure 44 (a-c): a(2014), b(2018), c(2022). It highlights the number of injuries tagged to each weather station over the year. For the EAT-based methods on the NSD dataset, two weights, $\alpha$ and $\beta$ were assigned to minority class samples with different accident types (high fatality risk, low fatality risk), such $\alpha + \beta = 1$.

The dataset is imbalanced (class unbalance ratio between major and minor injuries: $\frac{575}{10182} \approx \frac{1}{18}$). Thus, without oversampling, the baseline balanced accuracy was low (blue line) at about 51.0%. Nevertheless, in Figure 4 (d), classification models based on oversampled data using our EAT-ROS, EAT-SMOTE and EAT-ADASYN performed better than using the standard oversampling methods on most $\alpha$ values. Balanced accuracies were 74.3%, 77.0% and 81.8% based on data oversampled by ROS, ADASYN and SMOTE, respectively. In Figure 4 (d), balanced accuracies are were higher with EAT-ROS (82.2%); EAT-ADASYN (81.2%) and EAT-SMOTE (84.0%). Our methods achieve an improvement of 2.2%-7.9%. In Figure 4 (e), EAT-based oversampling methods resulted in higher average balanced accuracy than other methods for all supervised learning algorithms used.

In summary, our methods based on extended accident triangle weights can effectively improve performance. From a practical standpoint, adding the extended accident triangle weights is easy and effective in boosting prediction performance.

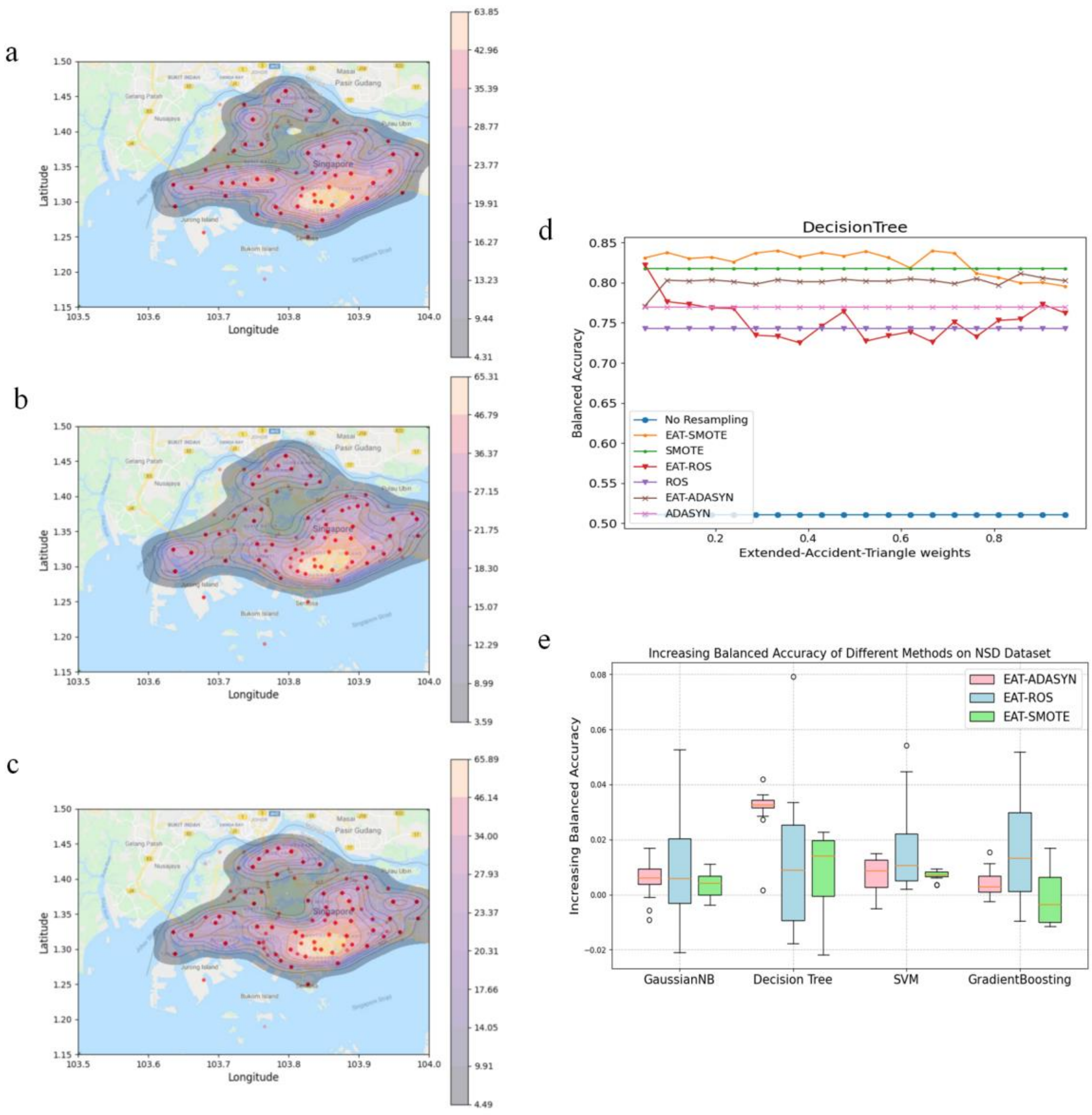

*Figure 4 Accident heatmaps in Singapore and performance comparison of our methods on NSD dataset. (a)-(c) The visualization of accident heatmaps in Singapore in 2014,2018,2022 (red points represent weather stations). (d) Comparsion of extended-accident-triangle series methods against baselines on Dataset NSD using Decision Tree. (e) The Increasing balanced accuracy of different methods on the NSD dataset.*

# 5 Discussion

## 5.1 Optimal extended-accident-triangle weights

In the experiments, we applied our EAT approach to different datasets and ML methods. To achieve the highest balanced accuracy, we searched for the optimal extended-accident-triangle weights for each dataset and method. We tried 20 different combinations of $\alpha$ and $\beta$, with $\alpha$ increasing from 0 to 1 by increments of 0.048 and $\beta$ decreasing from 1 to 0, correspondingly. Table 5 shows the weight pairs of $\alpha$ and $\beta$ that resulted in optimal classification model performance for different datasets, different classification algorithms and different oversampling methods. Note that some cells are blank in the table because all weights resulted in similar model performance for those combinations of oversampling and classification methods. It may be seen that the weight pairs that appear the most frequently include 0.952-0.048, 0.048-0.952, 0.238-0.762, 0.286-0.714, and 0.524-0.476. We recommend these weights when researchers use EAT-based oversampling to train ML and DL models. One may start by using the $\alpha$ value of around 0.95 or 0.05. If the resulting classification accuracy is not satisfactory, researchers may subsequently use $\alpha$ value around 0.25 and near 0.5.

Table 5 *Optimal extended-accident-triangle weights. '-' means that there is a small difference between increased balanced accuracies using different α and β. It means that we can choose many αs to obtain similar performance.*

| Dataset | Method | EAT-ROS | | EAT-SMOTE | | EAT-ADASYN | |
|---|---|---|---|---|---|---|---|
| | | $\alpha$ | $\beta$ | $\alpha$ | $\beta$ | $\alpha$ | $\beta$ |
| SMD | GaussianNB | 0.381 | 0.619 | 0.048 | 0.952 | - | - |
| | SMD Decision Tree | 0.333 | 0.666 | - | - | 0.429 | 0.571 |
| | Logistic Regression | 0.286 | 0.714 | 0.143 | 0.857 | 0.048 | 0.952 |
| | Convolutional Neural Network | 0.952 | 0.048 | 0.238 | 0.762 | 0.048 | 0.952 |
| | GaussianNB | 0.952 | 0.048 | 0.524 | 0.476 | 0.429 | 0.571 |
| SCD | SCD Logistic Regression | 0.952 | 0.048 | - | - | - | - |
| | Decision Tree | 0.952 | 0.048 | - | - | 0.762 | 0.238 |
| | GaussianNB | 0.095 | 0.904 | 0.095 | 0.904 | 0.524 | 0.476 |
| NSD | NSD Decision Tree | 0.048 | 0.952 | 0.095 | 0.904 | - | - |
| | GradientBoosting | 0.048 | 0.952 | 0.381 | 0.619 | - | - |
| | SVM | 0.048 | 0.952 | - | - | 0.619 | 0.381 |

## 5.2 SMD vs. SCD vs. NSD

The three datasets of this study reflect different levels of society and diverse aspects of workplace safety in Table 6. The data was collected at the national, industry and company levels, which are consistent with Rasmussen's (1997) understanding that the control of safety may occur at multiple levels of a sociotechnical system. Different stressors and safety controls act on different levels of a sociotechnical system. Thus, each level requires different considerations. For example, legal and political considerations are the most salient for safety control at the national level, while managerial and psychological considerations are the more salient at the company level. Testing the EAT method on the three datasets relevant to different levels of a sociotechnical system improves the validity of the evaluation. Similarly, the three datasets also represent different types of features, e.g., accident characteristics, company safety management data, and worker perception data, used in predictive safety analytics. This further underscores the versatility of the proposed methods.

*Table 6 Different levels and types of the three datasets. NSD: National Safety Dataset; SMD: Safety Management Dataset; SCD: Safety Climate Dataset*

| | **Data Type** | | |
|---|---|---|---|
| **Societal level** | **Mandatory accident report** | **Company safety management data** | **Safety climate perception survey** |
| **National** | NSD | | |
| **Industry** | NSD | | SCD |
| **Company** | | SMD | SCD |

## 5.3 EAT-ROS vs. EAT-SMOTE vs. EAT-ADASYN

We also compared the change in accuracy when oversampling different datasets using the EAT-based methods to see which EAT-based method is most suited to which dataset. In Figure 3 for the SMD dataset, the mean increase in accuracy was the highest for EAT-ADASYN for three out of four classification algorithms. This means EAT-ADASYN is the most suitable oversampling method for SMD. For SCD, EAT-SMOTE resulted in the highest mean accuracy increase for all classification algorithms tested. Thus, EAT-SMOTE is the most suitable for SCD. For NSD, EAT-ADASYN resulted in the highest mean accuracy increase for all classification algorithms tested. Thus, EAT-ADASYN is the most suitable for NSD.

## 5.4 Interpretations of the accident triangle

In Section 1 and Figure 2, the extended accident triangles have been established, demonstrating a layered relation between accidents at different levels. The key point is that the importance of each sample in the minority class may be treated differently based on a variable identified through domain knowledge. A weight can be assigned to each sample to adjust its possibility of being utilized in oversampling. Note that the weight of the accident triangles may differ from dataset to dataset.

As highlighted in Section 1, there are two possible interpretations of EAT when oversampling: give more emphasis to low-frequency but high-impact incidents or more emphasis on high-frequency low-impact accidents. It means that higher weights may be assigned to higher frequency accidents or higher impact accidents. Previous studies have not explored how to interpret the implications of the accident triangle to facilitate the selection of sampling weights. We compare them in Table 7.

For each EAT-based method and dataset, we summarise the balanced accuracies, conditioning on two possible EAT weight assignments. For EAT-SMOTE and EAT-ADASYN, we can see that emphasising high-impact, low-frequency accidents can produce models with better accuracy. However, interestingly, for SMD and NSD, EAT-ROS methods pay more attention to low-impact, high-frequency accidents, which supports the original accident triangle theory. Moreover, the total number of (b) in Table 7 is bigger than the number of (a). The results highlight the importance of high-impact, low-frequency accidents when oversampling.

Overall, the results imply that the dataset, oversampling method, and algorithm can influence the choice of weight. For example, EAT-SMOTE is less sensitive to the EAT weight, except when it is applied to NSD, where high-impact, low-frequency accidents should be given higher weight. Another example is EAT-ROS, which is linked to Table 6. In Table 6., in the column of EAT-ROS on the SCD dataset, the optimal EAT weights lie in the (0.952,0.048), which verifies case (b): we should assign the higher weight to high-impact (but low frequency) accidents. In summary, our results in Table 7 prefer the case b: high-impact, low-frequency accidents should be given higher weight, but it is not suitable for all situations. Thus, the results of this study can serve as a useful reference to future safety analytics researchers and practitioners.

*Table 7 Comparison of different EATs on three datasets. (a) means that the balanced accuracy becomes higher when we assign the higher weight to low-impact (but high frequency) accidents; (b) means that the balanced accuracy becomes higher when we assign the higher weight to high-impact (but low frequency) accidents; (≈) means that the balanced accuracies are similar when we assign the higher weight to two types of accidents.*

| **Dataset** | **Task** | **Method** | **EAT-ROS** | **EAT-SMOTE** | **EAT-ADASYN** |
|---|---|---|---|---|---|
| SMD | Occurrence prediction | GaussianNB | a | ≈ | ≈ |
| | | Decision Tree | a | ≈ | b |
| | | Logistic Regression | a | b | b |
| | | Convolutional Neural Network | b | ≈ | b |
| NSD | Severity prediction | Decision Tree | a | b | ≈ |
| | | GradientBoosting | b | b | ≈ |
| | | SVM | a | ≈ | b |
| | | GaussianNB | a | ≈ | ≈ |
| SCD | Occurrence prediction | GaussianNB | b | ≈ | ≈ |
| | | Logistic Regression | b | ≈ | ≈ |
| | | Decision Tree | b | ≈ | ≈ |

# 6 Limitations and Future Work

In this study, we only applied our insight into three oversampling methods. In fact, our general insight can be applied to other oversampling methods, following the same principle.

On the other hand, we released three safety-related datasets, and only tested our methods on these datasets. To facilitate the analyses on SCD and NSD, we used oversampling to create synthetic data, which may not be authentic in practical applications. However, in our analysis on SMD, we used actual data without synthetic data. Thus, future studies may investigate the suitability of our insights to ML-based safety prediction using larger and more practical data.

# 7 Conclusion

Safety analytics, a key tool to support safety management, leverages advanced quantitative methods to prevent workplace accidents. Its importance is especially prominent in high risk industries such as construction and trucking. However, imbalanced data is a severe challenge in safety analytics. To promote the development of safety analytics, we establish three imbalanced datasets related to accident prediction: a 9-year Singapore-wide construction accident record dataset, a Singapore construction safety management dataset, and an US trucking safety climate survey dataset. By introducing domain knowledge (the accident triangle), we improve the mainstream oversamplers. Because our insight is general, it can be applied to different oversamplers. We conducted practical experiments on three datasets, using different machine learning (GaussianNB, DT, SVM) and deep learning (neural network) methods. With the guidance of accident triangle weights, experimental results achieve superior performance and demonstrate the effectiveness of our insight. We provided the recommended extended-accident-triangle weights for researchers to train their ML models. The code is available at https://github.com/NUS-DBE/accident.

Our datasets and insight are representative of different safety datasets; this study has the potential to be applied in safety analytics tasks. We hope our datasets and insights can improve the use of safety analytics in academia and industry.

[The proposed datasets are available for reviewers and editors first. After the peer-review, the proposed datasets will be freely available to any researcher wishing to use them for non-commercial purposes.]

## Acknowledgements

SCD was collected while one of the authors worked at Liberty Mutual Research Institute for Safety. We thank the following team members for their invaluable assistance: Michelle Robertson, Susan Jeffries, Peg Rothwell, and Angela Garabet for data collection, analysis, and general assistance. We thank the Singapore Ministry of Manpower for supplying NSD and approving it for sharing. Finally, we would like to acknowledge the support from the anonymous construction organization which supplied SMD and approved it for sharing.